\documentclass[conference,a4paper]{APSIPA2012}
\usepackage{multirow}
\usepackage[dvips]{graphicx}
\usepackage{amsmath}
\usepackage[psamsfonts]{amssymb}
\usepackage{amsxtra}
\usepackage{graphicx,color}
\usepackage{threeparttable}
\usepackage{url}
\usepackage{epsfig,epstopdf}

\begin{document}

\title{AP16-OL7: A Multilingual Database for Oriental Languages and A Language Recognition Baseline}

\author{%
\authorblockN{%
Dong Wang\authorrefmark{1},
Lantian Li\authorrefmark{1},
Difei Tang\authorrefmark{2} and
Qing Chen\authorrefmark{2}
}
\authorblockA{%
\authorrefmark{1}
Center for Speech and Language Technologies, Division of Technical Innovation and Development, \\
Tsinghua National Laboratory for Information Science and Technology;\\
Center for Speech and Language Technologies, Research Institute of Information Technology;\\
Department of Computer Science and Technology, Tsinghua University, Beijing, China\\
E-mail: \{wangdong99,lilt13\}@mails.tsinghua.edu.cn}
\authorblockA{%
\authorrefmark{2}
SpeechOcean\\
E-mail:\{tangdifei,chenqing\}@speechocean.com}
}

\maketitle
\thispagestyle{empty}

\begin{abstract}

We present the AP16-OL7 database which was released as the training and test data
for the oriental language recognition (OLR) challenge on APSIPA 2016. Based on the
database, a baseline system was constructed on the basis of the i-vector model. We report the
baseline results evaluated in various metrics defined by the AP16-OLR evaluation plan
and demonstrate that AP16-OL7 is a reasonable data resource for multilingual research.

\end{abstract}

\section{Introduction}

Oriental languages, including various languages spoken in east, northeast and southeast Asia, belong to
several language families, including Austroasiatic languages (e.g.,Vietnamese, Cambodia )~\cite{sidwell201114},
Tai–Kadai languages (e.g., Thai, Lao), Hmong–Mien languages (e.g., some dialects in south China), Sino-Tibetan languages (e.g., Chinese Mandarin), Altaic languages (e.g., Korea, Japanese), Indo-European languages (e.g., Russian)~\cite{ramsey1987languages,shibatani1990languages,comrie1996russian}. These languages were generally
believed to be genetically unrelated and were developed from diverse cultures. However, they do share many features
due to the demographic migration and international business interaction in history.
For example, many
languages in the so-called Mainland Southeast Asia (MSEA) linguistic area posses
a particular syllable structure that involves monosyllabic morphemes, lexical tone, a fairly large inventory of
consonants~\cite{enfield2005areal}. Another example is the significant influence of
Chinese to Korean, Japanese, Vietnamese and many languages in southeast Asia. In the modern period,
English becomes the most influential language, resulting in numerous English-originated words
in almost all oriental languages.

The complex acoustic and linguistic patterns of oriental languages
have attracted much interest in a multitude of research areas, including
comparative phonetics, evolutionary linguistics, second language acquisition, and social linguistics.
In particular, the diverse evolution paths of these languages and their complicated
interaction offers a valuable opportunity for studying mixlingual and multilingual phenomena.

Despite the broad interest, data resources of oriental languages are far from abundant. One
possible reason is that many of these languages are spoken by a relatively small population,
and most of the speakers are in developing countries.
Some effort has been devoted to building data resources for oriental languages, e.g.,
the annual oriental COCOSDA (OC)
workshop intends to promote speech and
language resource construction for oriental languages, and the transactions on Asian and Low-Resource Language
Information Processing (TALLIP) journal calls for original research on oriental languages, especially
languages with limited resources.\footnote{\url{https://mc.manuscriptcentral.com/tallip}} Some projects,
e.g., the Babel program\footnote{\url{https://www.iarpa.gov/index.php/research-programs/babel}},
although not particularly for oriental languages, do involve Vietnamese, Thais, Lao and
some other low-resource languages in southeast Asia. In spite of these efforts, resource construction and corresponding
research on oriental languages are still rather limited, except one or two rich-resource languages,
such as Chinese and Japanese.

To promote research for oriental languages, particularly on multilingual speech and language processing,
the center for speech and language technologies (CSLT) at Tsinghua
University and Speechocean collaborated together and organized an oriental language recognition (OLR) challenge on
APSIPA 2016. This event called for a competition on a language recognition task on seven oriental languages.
To support this event, Speechocean released a multilingual speech database AP16-OL7 and made it free for the
challenge participants. This paper will present the data profile of the database, the evaluation rules of the
challenge, and a baseline system that the participants can refer to.

Note that there are several databases that can be used for multilingual research.
For example, polyphone~\cite{godfrey1994multilingual},
globalPhone~\cite{schultz2002globalphone}, NTT multilingual database\footnote{\url{http://www.ntt-at.com/product/speech2002/}},
SPEECHDAT-CAR~\cite{moreno2000speechdat},Speechdat-E~\cite{van2001speechdat}, Babel~\cite{roach1996babel}, and
the multilingual databases created by the new Babel project. To our best knowledge, AP16-OL7 is the first
multilingual speech database specifically designed for oriental languages.

\section{Database profile}

\begin{table*}
\begin{center}
\caption{AP16-OL7 Data Profile}
\label{tab:ol7}
\begin{tabular}{|l|l|c|c|c|c|c|c|c|}
\hline
\multicolumn{3}{|c|}{Datasets} & \multicolumn{3}{c|}{Training \& Dev set}  & \multicolumn{3}{c|}{Test set}\\
\hline
Code & Description & Channel & No. of Speakers & Utt./Spk. & Total Utt. & No. of Speakers & Utt./Spk. & Total Utt. \\
\hline
ct-cn & Cantonese in China Mainland and Hongkong & Mobile & 18 & 320 & 5759 & 6 & 320 & 1920 \\
\hline
zh-cn & Mandarin in China & Mobile & 18 & 300 & 5398 & 6 & 300 & 1800 \\
\hline
id-id & Indonesian in Indonesia &  Mobile & 18 & 320 & 5751 & 6 & 320 & 1920 \\
\hline
ja-jp & Japanese in Japan & Mobile & 18 & 320 & 5742 & 6 & 320 & 1920 \\
\hline
ru-ru & Russian in Russia & Mobile & 18 & 300 & 5390 & 6 & 300 & 1800 \\
\hline
ko-kr & Korean in Korea & Mobile & 18 & 300 & 5396 & 6 & 300 & 1800 \\
\hline
vi-vn & Vietnamese in Vietnam & Mobile & 18 & 300 & 5400 & 6 & 300 & 1800 \\
 \hline
\end{tabular}
\begin{tablenotes}
\item[a] Male and Female speakers are balanced.
\item[b] The number of total utterances might be slightly smaller than expected, due to the quality check.
\end{tablenotes}
\end{center}
\end{table*}

The AP16-OL7 database was originally created by Speechocean targeting for various speech processing tasks (mainly
speech recognition). The entire database involves seven datasets, each in a particular language. The seven languages are:
Mandarin, Cantonese, Indoesian, Japanese, Russian, Korean, Vietnamese.
The data volume for each language is about $10$ hours of speech signals recorded by $24$
speakers ($12$ males and $12$ females), and each speaker recorded about $300$ utterances in
reading style. The signals were
recorded by mobile phones, with a sampling rate of $16$kHz  and a sample size of $16$ bits.
Each dataset
was split into a training set consisting of $18$
speakers, and a test set consisting of $6$ speakers.
For Mandarin, Cantonese, Vietnamese and Indonesia, the recording was conducted in a quiet environment.
As for Russian, Korean and Japanese, there are $2$ recording sessions for each speaker: the first session
was recorded in a quiet environment and the second was recorded in a noisy environment.
The basic information of the AP16-OL7 database is presented in Table~\ref{tab:ol7}.

Besides the speech signals, the AP16-OL7 database also provides lexicons of all the seven languages, and
transcriptions of all the training utterances. These resources allow training acoustic-based or phonetic-based
language recognition systems. Training phone-based speech recognition systems is also possible, though
large vocabulary recognition systems are not well supported, due to the lack of large-scale language models.

The AP16-OL7 database is freely available for the participants of the AP16-OLR challenge and the APSIPA 2016 special
session on {\it multilingual speech and language processing}. It is also available for any academic and industrial
users, subject to a slightly different licence from SpeechOcean.\footnote{\url{http://speechocean.com}}

\section{AP16-OLR challenge}

Based on the AP16-OL7 database, we call an oriental language recognition (OLR)
challenge.\footnote{\url{http://cslt.riit.tsinghua.edu.cn/mediawiki/index.php/ASR-events-AP16-details}}
Following the definition of NIST LRE15~\cite{lre15}, the task of the challenge is defined
as follows: Given  a  segment  of  speech  and  a  language  hypothesis (i.e.,  a  target
language  of  interest  to  be  detected),  the  task  is  to decide  whether  that
target  language  was  in  fact  spoken  in  the given segment (yes or no), based on an
automated an analysis of the data contained in the segment.
The AP16-OLR evaluation plan also follows the principles of NIST LRE15: it focuses on
the close-set condition, and allows no additional training materials besides AP16-OL7.
The evaluation details are described as follows.

\subsection{System input/output}

The input to the OLR system is a set of speech segments in unknown languages (but within
the $7$ languages of AP16-OL7). The task of the OLR system is to determine
the confidence that a language is contained in a speech segment. More specifically,
for each speech segment, the OLR system outputs a score vector $<\ell_1, \ell_2, ..., \ell_7>$,
where $\ell_i$ represents the confidence that language $i$ is spoken in the speech segment.
Each score $\ell_i$ will be interpreted as follows: if $\ell_i \ge 0$, then the decision
would be that language $i$ is contained in the segment, otherwise it is not. The scores
should be comparable across languages and segments.
This is consistent with
the principle of LRE15, but differs from that of LRE09~\cite{lre09} where an explicit decision
is required for each trial.

In summary, the output of an OLR submission will be a text file, where each line contains
a speech segment plus a score vector for this segment, e.g.,

\vspace{0.5cm}
\begin{tabular}{cccccccc}
seg$_1$ & 0.5  & -0.2 & -0.3 & 0.1 & -9.2 & -0.1 & -5.1 \\
seg$_2$ & -0.1 & -0.3 & 0.5  & 0.3 & -0.5 & -0.9 & -3.2 \\
...   &      &      &      &  ...   &      &      &
\end{tabular}

\subsection{Test condition}


\begin{itemize}
\item No additional training materials are allowed to use.
\item All the trials should be processed. Scores of lost trials will be interpreted as -$\inf$.
\item Each test segment should be processed independently. Knowledge from other test segments is not allowed to use (e.g.,
score distribution of all the test segments).
\item Information of speakers is not allowed to use.
\item Listening to any speech segments is not allowed.
\end{itemize}

\subsection{Evaluation metrics}

As in LRE15, the AP16-OLR challenge chooses $C_{avg}$ as the principle evaluation metric.
First define the pair-wise loss that composes the missing and
false alarm probabilities for a particular target/non-target language pair:

\[
C(L_t, L_n)=P_{Target} P_{Miss}(L_t) + (1-P_{Target}) P_{FA}(L_t, L_n)
\]

\noindent where $L_t$ and $L_n$ are the target and non-target languages, respectively; $P_{Miss}$ and
$P_{FA}$ are the missing and false alarm probabilities, respectively. $P_{target}$ is the prior
probability for the target language, which is set to $0.5$ in the evaluation. Then the principle metric
$C_{avg}$ is defined as the average of the above pair-wise performance:


\[
 C_{avg} = \frac{1}{N} \sum_{L_t} \left\{
\begin{aligned}
  & \ P_{Target} \cdot P_{Miss}(L_t) \\
  &  + \sum_{L_n}\ P_{Non-Target} \cdot P_{FA}(L_t, L_n)\
\end{aligned}
\right\}
\]

\noindent where $N$ is the number of languages, and $P_{Non-Target}$ = $(1-P_{Target}) / (N -1 )$.

\section{Baseline results}

We present baseline language recognition systems based on the i-vector model, and evaluate the performance
in terms of the metrics defined by the AP16-OLR challenge.
The purpose of these experiments is not to present a competitive submission, instead
to demonstrate that the AP16-OL7 database is a reasonable data resource to conduct
language recognition research.

%

\subsection{Experimental setup}

The baseline system was constructed based on the i-vector model~\cite{dehak2011front-end,dehak2011language}.
The static acoustic features involved 19-dimensional Mel
frequency cepstral coefficients (MFCCs) and the log energy.
This static features were augmented by their first and second order derivatives, resulting in 60-dimensional
feature vectors.
The UBM involved $2,048$ Gaussian components and the dimensionality of the i-vectors was $400$.
Linear discriminative analysis (LDA) was employed to promote language-related information.
The dimensionality of the LDA projection space was set to $6$.

With the i-vectors (either original or after LDA transform), the score of a trail on a particular language
can be simply computed as the cosine distance between the test i-vector and the mean i-vector of
the training segments that belong to that language. This is denoted to
be `cosine distance scoring'.
A more powerful
scoring approach is to employ various discriminative models. In our experiment, we
trained a support vector machine (SVM)  for each language to determine the score
that a test i-vector belongs to that language. The SVMs were trained on the i-vectors
of all the training segments, following the one-verse-rest scheme. We will call this scoring
approach as `SVM-based scoring'.

\subsection{Visualization with T-SNE~\cite{saaten2008}}

\begin{figure}[htb]
\begin{center}
\includegraphics[width=0.95\linewidth]{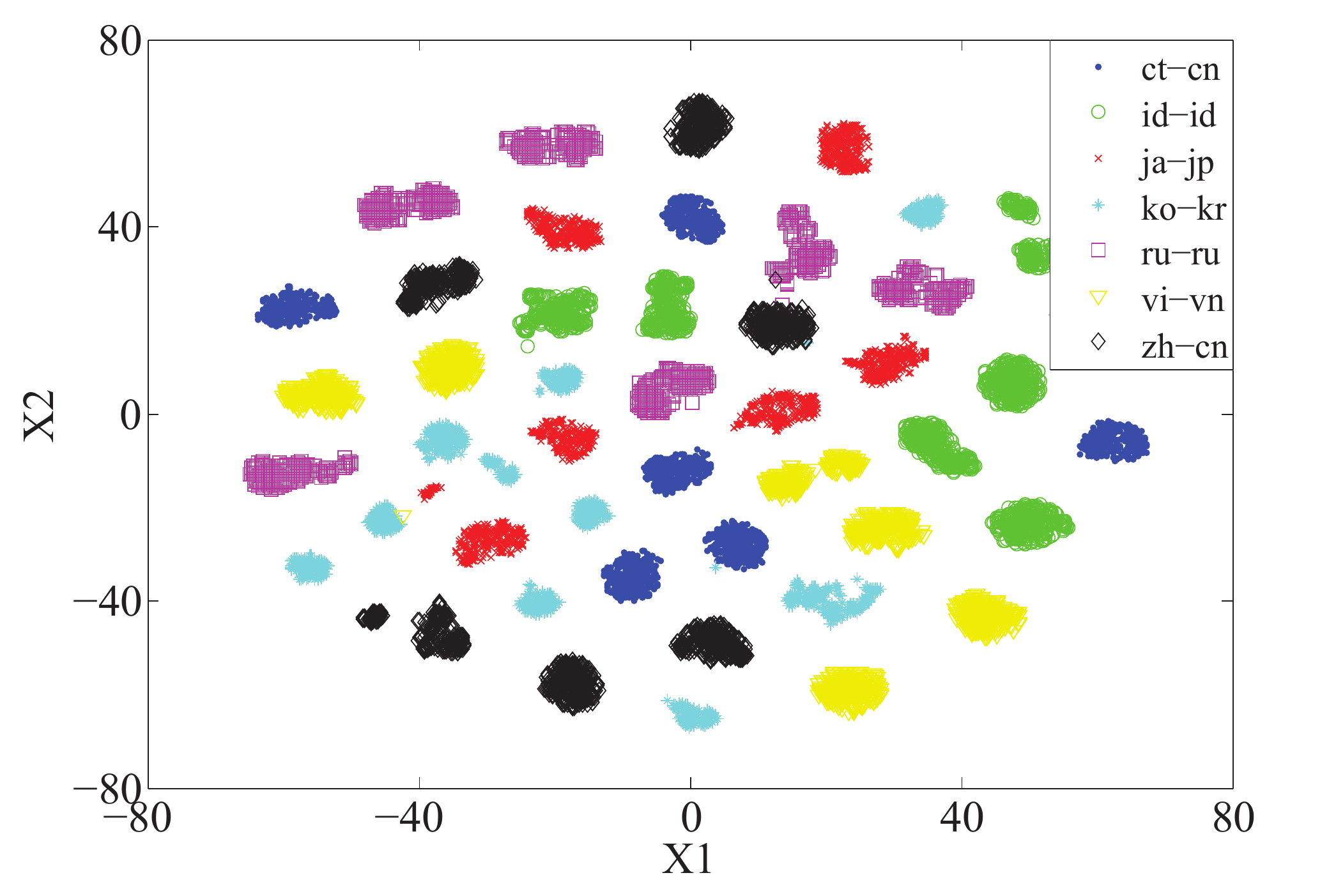}
\end{center}
\caption{Original i-vectors plotted by t-SNE. Each color/shape represents a particular language.}
\label{fig:tsne-1}
\end{figure}

\begin{figure}[htb]
\begin{center}
\includegraphics[width=0.95\linewidth]{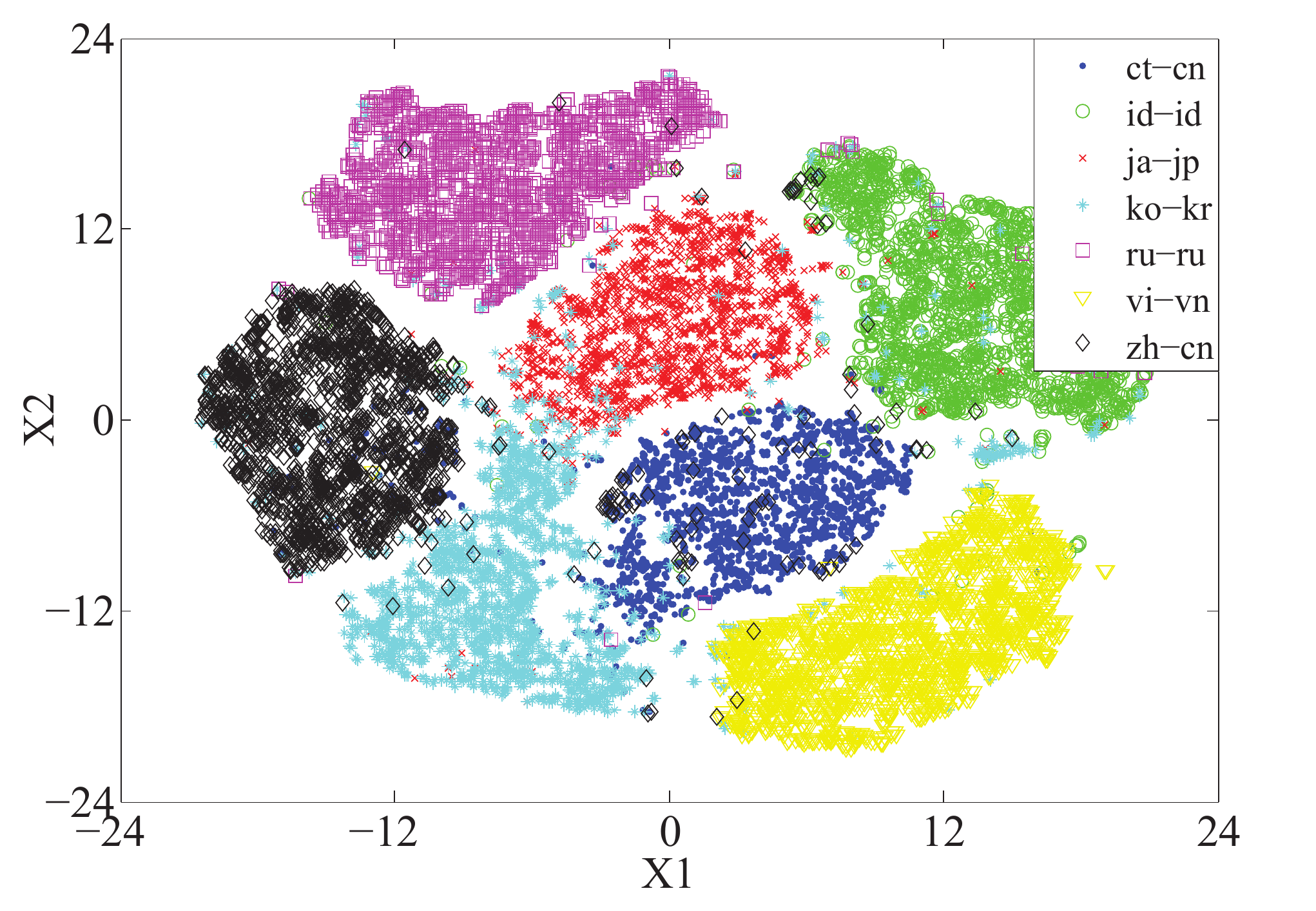}
\end{center}
\caption{LDA-transformed i-vectors plotted by t-SNE. Each color/shape represents a particular language.}
\label{fig:tsne-2}
\end{figure}

To provide an intuitive understanding of the discriminative capability of i-vectors on languages,
the i-vectors of all the segments in the test set are plotted in a two-dimensional space via
T-SNE~\cite{saaten2008}. Fig. ~\ref{fig:tsne-1} shows the original i-vectors, and Fig. ~\ref{fig:tsne-2} shows the
i-vectors after LDA transform, where each color/shap represents a particular language. It can be seen that
for the original i-vectors, each language is split into several clusters basically due to
different speakers. After LDA transformation, speaker information is suppressed and the language identify
is more significant.

\subsection{Performance results}

The primary evaluation metric in AP16-OLR is $C_{avg}$. Besides that, we also present the performance
in terms of equal error rate (EER), minimum detection cost function ($min$DCF),
detection error tradeoff (DET) curve, and identification rate (IDR). These metrics evaluate the
system from different perspectives, offering a whole picture of the verification/identification capability
of the baseline system.

\subsubsection{$C_{avg}$ results}

The $C_{avg}$ results are shown in Table~\ref{tab:result}.
The rows `i-vector' and `L-vector' present the results with the cosine distance scoring; `i-vector-SVM' and
`L-vecotr-SVM' present the results with the SVM-based scoring. `Linear', `Poly'(degree=$3$), and `RBF' represent the three
commonly used kernel functions. It can be seen that LDA leads to consistent performance gains, and the SVM-based
scoring tends to outperform cosine distance scoring.

\subsubsection{EER and $min$DCF results}

EER and $min$DCF are also widely used in measuring performance of verification systems.
Compared to $C_{avg}$,
these two metrics are not related to the decision result, but the quality of the scoring, and therefore
evaluate the verification system from a different angle. The results for these two metrics
are presented in Table~\ref{tab:result}.
respectively. It can be seen that similar conclusions can be drawn from these results as from
the $C_{avg}$ results.

\begin{table}[htb]
\normalsize
\begin{center}
\caption{C$_{avg}$, EER, $min$DCF $and$ IDR results of various baseline systems}
\label{tab:result}
\begin{tabular}{|l|c|c|c|c|}
\hline
System        &  $C_{avg}$*100  &   EER\%  & $min$DCF  & IDR\% \\
\hline
\hline
i-vector      &  5.63           &   6.65    &  0.0659   &  89.16 \\
L-vector      &  4.15           &   4.76    &  0.0472   &  90.19 \\
\hline
\hline
i-vector-SVM  &  5.68           &   5.62    &  0.0558   &  87.07 \\
(Linear)      &                 &           &           &     \\
i-vector-SVM  &  3.06           &   3.06    &  0.0303   &  92.73  \\
(Poly)        &                 &           &           &     \\
i-vector-SVM  &  3.86           &   3.83    &  0.0381   &  90.80  \\
(RBF)         &                 &           &           &     \\
\hline
\hline
L-vector-SVM  &  3.52           &   3.49    &  0.0344   &  91.82  \\
(Linear)      &                 &           &           &     \\
L-vector-SVM  &  3.37           &   3.37    &  0.0334   &  91.99  \\
(Poly)        &                 &           &           &     \\
L-vector-SVM  &  3.40           &   3.36    &  0.0333   &  92.04  \\
(RBF)         &                 &           &           &     \\
\hline
\end{tabular}
\end{center}
\end{table}

\subsection{DET curve}

\begin{figure}[htb]
\begin{center}
\includegraphics[width=0.95\linewidth]{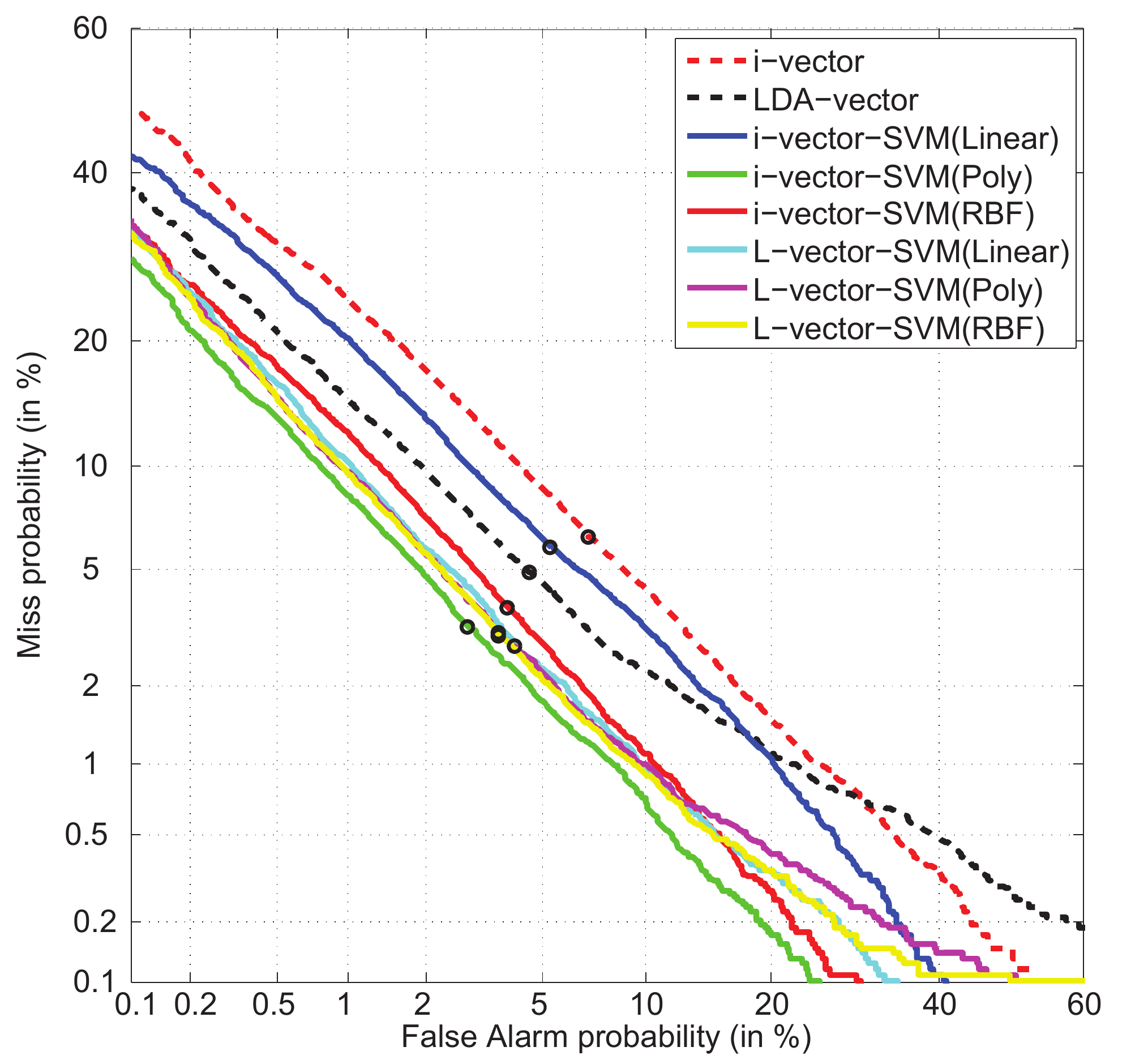}
\end{center}
\caption{The DET curves of various baseline systems.}
\label{fig:det-1}
\end{figure}

The DET curve is another popular way to evaluate verification systems. Compared to $C_{avg}$, EER and
$min$DCF, the DET curve presents performance on all operation points, and therefore can evaluate
a verification system in a more systematic way. Experimental results are shown in Fig~\ref{fig:det-1}.
The black circles represent the operation location where the $min$DCFs are obtained.
Again, similar conclusions as with the $C_{avg}$, EER and $min$DCF can be obtained.

\subsubsection{IDR results}

Note that in the OLR challenge, the target languages are known in prior, and the confidence scores are
comparable across languages. This means that OLR can be treated as a language identification task,
for which the language obtaining the highest score in a trail is regarded as the identification result.
For such an identification task, IDR is a widely used metric~\cite{yinbo2002lid}, which treats errors on
all languages equally serious. IDR is formally defined as follows:

\[
IDR = \frac {T_c} {T_c + T_i}
\]

\noindent where $T_c$ and $T_i$ are the numbers of correctly and incorrectly identified utterances,
respectively.
Table~\ref{tab:result} presents the IDR results of the baseline system. We can observe similar trends
as with the verification metrics: $C_{avg}$, EER, $min$DCF and DET curve.

\section{Conclusions}

We presented the data profile of the AP16-OL7 database that was released to support the
AP16-OLR challenge on APSIPA 2016. The evaluation rules of the challenge was described,
and a baseline system was presented. We show that the AP16-OL7 database is a suitable
data resource for language recognition research.

\section*{Acknowledgment}

This work was supported by the National Science Foundation of China (NSFC)
under the project No. 61371136, and the MESTDC PhD Foundation Project No.
20130002120011. It was also supported by SpeechOcean.

\bibliographystyle{IEEEtran}
\bibliography{ole}

\end{document}